\documentclass[letterpaper, 10 pt, conference]{ieeeconf}

\IEEEoverridecommandlockouts

\usepackage{graphics}
\usepackage{epsfig}
\usepackage{mathptmx}
\usepackage{times}
\usepackage{amsmath}
\usepackage{amssymb}

\usepackage{booktabs}

\usepackage{cleveref}
\usepackage{siunitx}
\usepackage{xcolor}
\definecolor{OIblack}{RGB}{0, 0, 0}
\definecolor{OIgreen}{RGB}{0, 158, 115}
\definecolor{OIblue}{RGB}{0, 114, 178}
\definecolor{OIlightblue}{RGB}{86, 180, 233}
\definecolor{OIyellow}{RGB}{240, 228, 66}
\definecolor{OIorange}{RGB}{230, 159, 0}
\definecolor{OIred}{RGB}{213, 94, 0}
\definecolor{OIpink}{RGB}{204, 121, 167}
\usepackage{tikz}
\usetikzlibrary{arrows.meta, shapes.geometric, positioning, calc, fit}
\usepackage{multirow}
\usepackage{makecell}
\usepackage{tabularx}
\usepackage[acronym]{glossaries}
\newacronym{dvrk}{dVRK}{da Vinci Research Kit}
\newacronym{mops}{MOPS}{A Modular and Open Platform for Surgical Robotics Research}

\title{\LARGE \bf
An Open-Source da Vinci X/Xi Instrument Module for Laparoscopic Surgery Research
}

\author{
Douglas McDonough$^{1,\dagger}$,
Justin Liao$^{1,\dagger}$,
Jiawei Ge$^{1}$,
Antony Goldenberg$^{1}$,
Yvonne Zhang$^{1}$, \\
Mackenzie A Eagleson$^{2}$,
Ravi Prakash$^{1}$,
Axel Krieger$^{1,\ddagger}$,
and Chunwoo Kim$^{3,\ddagger}$%
\thanks{*This work was supported by the Advanced Research Projects Agency for Health (ARPA-H) under grant number AY1AX000023, and the National Science Foundation (NSF/FRR CAREER) under grant number 2144348.}%
\thanks{$^{\dagger}$Co-first authors. $^{\ddagger}$Co-senior authors.}%
\thanks{$^{1}$Laboratory for Computational Sensing and Robotics, Johns Hopkins University, Baltimore, MD 21218, USA.}%
\thanks{$^{2}$Department of Surgery, Johns Hopkins Medicine, Baltimore, MD 21287, USA.}%
\thanks{$^{3}$Korea Institute of Science and Technology (KIST), Seoul 02792, Republic of Korea.}%
}

\begin{document}

\maketitle
\thispagestyle{empty}
\pagestyle{empty}

\begin{abstract}

Studying electrosurgical tasks on a research robot requires control of both instrument motion and energy activation, along with a record of when each action occurs. This paper presents an open-source module for da Vinci X/Xi instruments on a UR5/UR5e robotic arm. Five motors transmit torque through timing belts to spring-loaded drive discs. An electrosurgical interposer accepts software commands, gives the surgeon's foot pedals priority, and records the source of each activation request. In bench testing, the interposer recorded all 12 activation and deactivation events from six pedal presses. Ex vivo demonstrations included monopolar and bipolar coagulation of splenic tissue and monopolar cutting during gallbladder dissection. Optical tracking of the Fenestrated Bipolar Forceps showed that calibration brought pitch and yaw gains from 0.904 and 1.14 to 1.00 and 1.02, respectively. The module enables researchers to record instrument motion, video, and the timing and source of electrosurgical commands during laparoscopic tasks.
\end{abstract}

\section{INTRODUCTION}

\label{sec:introduction}

Numerous research platforms have been developed to accelerate research in robot-assisted surgery~\cite{hannaford2013raven, kazanzides2014dvrk, xu2025dvrksi, schwaner2021mops, rodriguez2026opensource, wickenhiser2026dash}. Unlike clinical systems, these platforms give researchers access to low-level hardware data and control, enabling the developments and evaluations of new control algorithms and autonomous functions.

Earlier platforms required developing large parts of the robotic system. The Raven-II~\cite{hannaford2013raven} was built almost entirely in-house, including its cable-driven arms, instruments~\cite{moreyra2005wrist}, control electronics, and software, and was later sold through the spin-off company Applied Dexterity~\cite{kazanzides2014dvrk}. The da Vinci Research Kit (dVRK)~\cite{kazanzides2014dvrk, xu2025dvrksi} instead reuses manipulators and instruments from retired da Vinci systems, typically donated by Intuitive Surgical, and provides open-source control electronics and software, with assembled controllers purchased separately. In both cases, acquiring the
hardware, whether by buying a complete system or by applying for donated equipment and buying controllers, has limited how widely these platforms can be adopted.
\begin{figure}[!h]
    \centering
    \includegraphics[height=6cm]{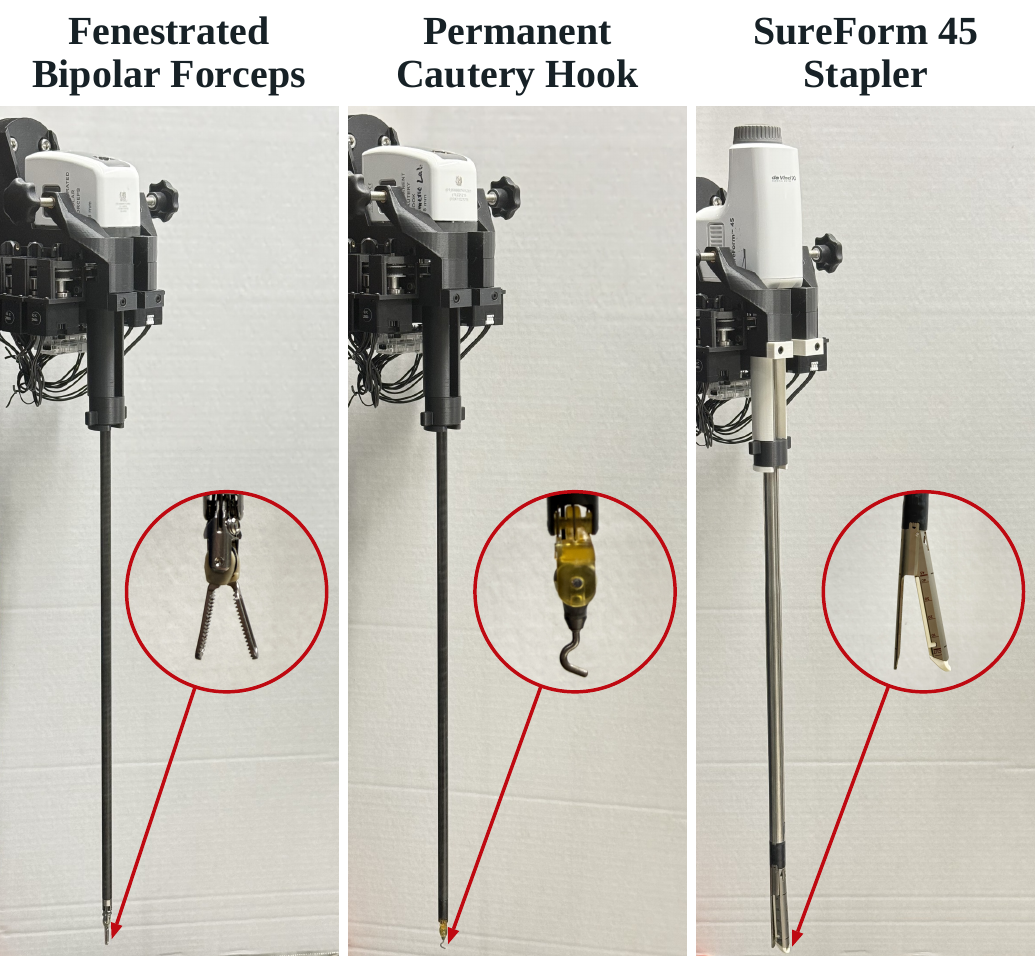}
    \caption{Fenestrated Bipolar Forceps, Permanent Cautery Hook, and SureForm 45 Stapler mounted on the instrument driver, from left to right. Circular insets show the corresponding end effectors.}
    \label{fig:instruments}
\end{figure}

More recent platforms reduce this burden by building on commercially
available robot arms, whose vendors supply the arm mechanism, control
electronics, and low-level control~\cite{schwaner2021mops,
rodriguez2026opensource, wickenhiser2026dash}. MOPS~\cite{schwaner2021mops} and DASH~\cite{wickenhiser2026dash} add easy-to-build adapters that actuate da Vinci EndoWrist instruments, taking advantage of the availability of used instruments and letting surgeons work with instruments they already know. With the growing role of artificial intelligence in surgical robotics, recent platforms also emphasize data collection: Rodriguez~\textit{et al.}~\cite{rodriguez2026opensource} provide a stack for demonstration recording and deployment of learned policies, and DASH automatically identifies inserted instruments to load their kinematic parameters~\cite{wickenhiser2026dash}.

However, these adapters actuate only standard articulated instruments such as graspers, needle drivers, and scissors. To our knowledge, none supports advanced instruments such as staplers, and none provided interfaces for remote control and data logging of the electrosurgical instruments.

Electrosurgical usage data have been collected in previous studies~\cite{oh2023crcd}~ \cite{weerasinghe2026midas}. CRCD ~\cite{oh2023crcd} records console pedal signals and includes an Arduino interface between the monopolar pedal and generator; MiDAS ~\cite{weerasinghe2026midas} uses pedal-mounted force sensors. Oh \textit{et al.}~\cite{oh2023dissection} also triggered monopolar and bipolar energy from software through an Arduino on the dVRK for automated dissection. However, none of these works describes separate surgeon and software source labels for energy activation or a documented interface for software activation with surgeon priority.

This paper presents an open-source instrument module that combines a five-motor driver for da Vinci X/Xi instruments with an electrosurgical interface. The driver accommodates wristed instruments and the SureForm 45 Stapler. The interface gives physical pedals priority over software commands and records the source of activation requests alongside robot data.

\section{INSTRUMENT MODULE DESIGN}
\label{sec:module}

\subsection{Design Goals}
The instrument module was designed to provide a compact package for actuating da Vinci X/Xi surgical instruments. These include wristed tools such as fenestrated bipolar forceps, monopolar curved scissors, large needle drivers, and cautery hooks, as well as more advanced instruments such as the SureForm 45 stapler. Building upon a previous adapter based on MOPS~\cite{schwaner2021mops}, the redesign focused on three objectives: (1) upgrading the actuation components to incorporate a fifth motor and enable current-based control, (2) simplifying instrument engagement, and (3) improving tool securing to maintain tool position. These requirements guided the mechanical and actuation designs described in the following sections. The module's CAD, firmware and driver are part of the ARTS repository (\texttt{https://github.com/MRE-Lab-UMD/arts}), under hardware/tool\_adapter/ and arts\_ros2\_ws/src/arts\_embedded/ arts\_dv\_tool\_driver/.

\subsection{Disc Drive}
The instrument module uses five DYNAMIXEL XC330-M288-T servo motors to actuate the input discs of the da Vinci X/Xi surgical instruments. The design augments the previous MOPS adapter ~\cite{schwaner2021mops} by replacing the DYNAMIXEL MX-28T actuators with these smaller sizes. The XC330-M288-T measures 20 x 34 x 26 mm, compared to 35.6 x 50.6 x 35.5 mm for the MX-28T. This 72\% reduction in actuator bounding-box volume makes room for a fifth motor. The mass of each actuator is also reduced from 77 g to 23 g. Additionally, the XC330 supports current-based control, enabling motor current measurements to be used during instrument homing.

\begin{figure}[!t]
    \centering
    \includegraphics[height=5.5cm]{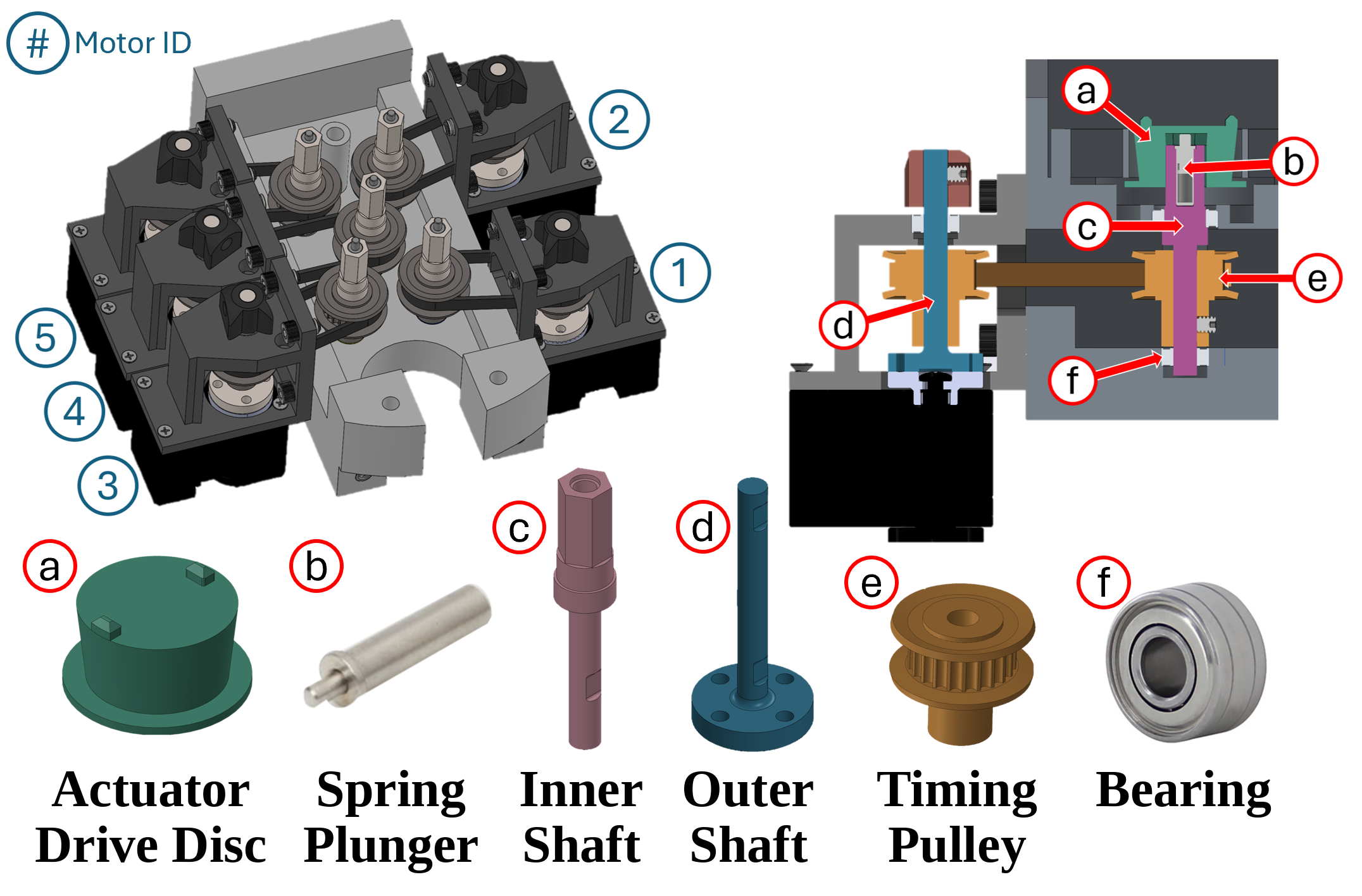}
    \caption{Five-motor instrument driver and cross-section of the belt-driven, spring-loaded disc assembly. Numbers identify the motors. Letters identify the components shown below the cross-section.}
    \label{fig:inners}
\end{figure}

Motor rotation is transmitted to the instrument drive discs through a compact timing-belt transmission consisting of 22-tooth pulleys, 106 mm and 128 mm timing belts, and ball bearings to support rotating shafts (Fig.~\ref{fig:inners}). This configuration allows the motors to be mounted on the exterior of the adapter while transmitting torque to their corresponding instrument drive disc. Two types of shafts were designed for machining. The first connects directly to the DYNAMIXEL motor horn and incorporates an externally accessible manual adjustment knob. Similar to MOPS~\cite{schwaner2021mops}, this redesigned feature allows users to manually engage the drive discs or reposition the instrument tip when the motors are in free-drive mode.

The second shaft is located within the adapter and is designed to facilitate instrument engagement. Each shaft holds a micro spring plunger with a 3 mm stroke, allowing axial translation of the adapter drive disc. This serves a similar function to the spring-loaded engagement mechanism in DASH~\cite{wickenhiser2026dash}. A hexagonal end connects the shaft to a custom-designed adapter drive disc, transmitting torque while permitting axial displacement. D-profile cuts provide flat surfaces for pulley set-screw connections, with threadlocker applied to prevent loosening during use. During instrument engagement, contact between the adapter and instrument discs compresses the spring plungers. As the adapter discs rotate, their prongs align with the corresponding recesses on the instrument discs, allowing the spring restoring force to push the adapter discs into engagement.

\begin{figure}[!t]
    \centering
    \includegraphics[height=5.0cm]{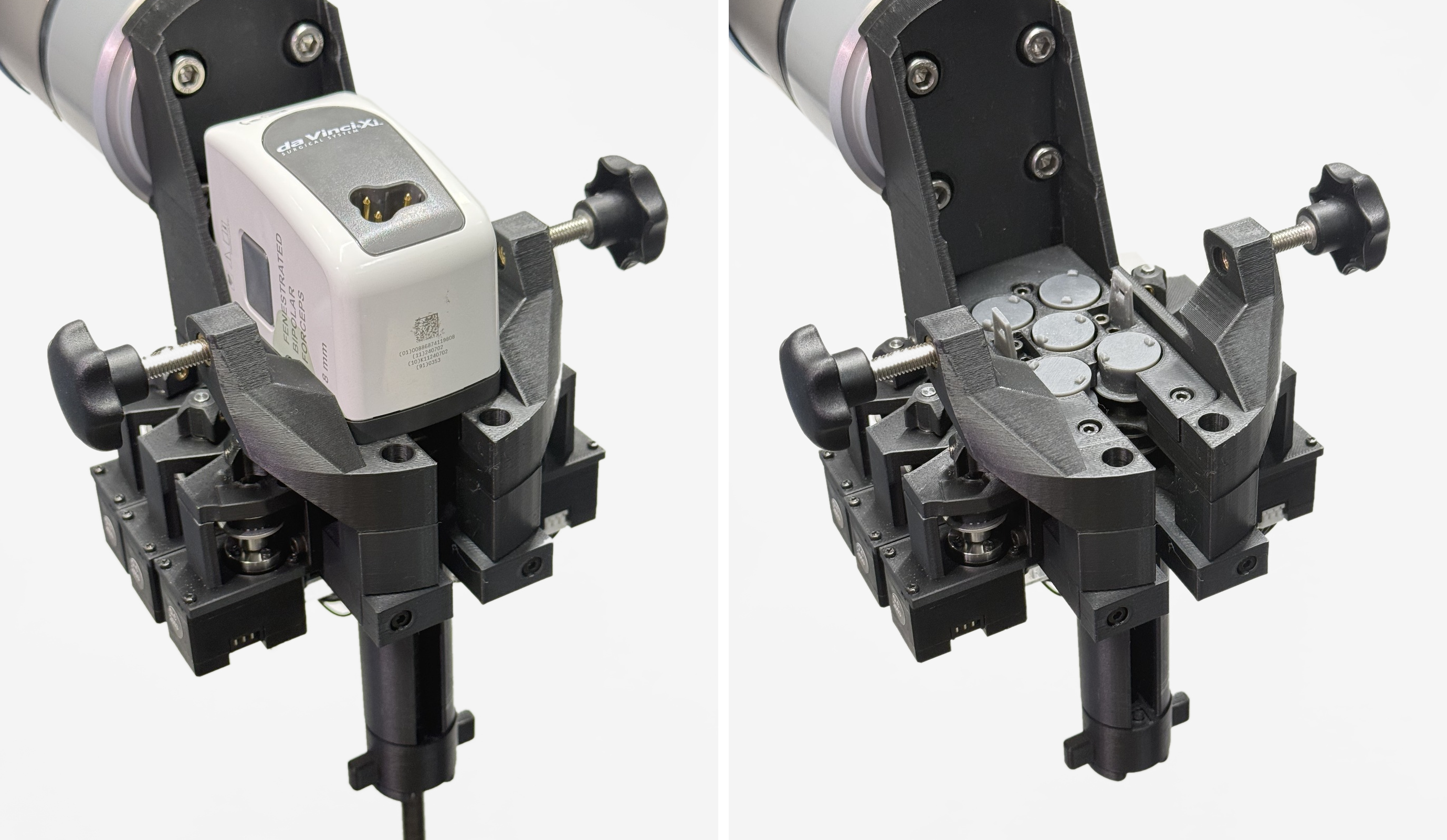}
    \caption{Instrument driver with the Fenestrated Bipolar Forceps engaged (left) and with the instrument removed to expose the five drive discs (right).}
    \label{fig:instrument_engagement}
\end{figure}

\subsection{Mounting}

The instrument module mounts to the UR5/UR5e robotic manipulator through an ISO 9409-1-50-4-M6 flange interface adapted from MOPS~\cite{schwaner2021mops}. An additional locating pin was added to the rear of the plate, engaging with a corresponding slot on the UR5 flange to ensure rotational alignment and prevent any rotation during use.

The da Vinci instrument is secured to the adapter using snap-fit clips that engage the instrument housing during insertion (Fig.~\ref{fig:instrument_engagement}), similar to the latching mechanism presented in DASH~\cite{wickenhiser2026dash}. The clips are the primary mechanism for securing the tool, allowing instruments to be inserted and removed without additional fasteners. Two manually adjustable star knobs provide an optional secondary securing mechanism for additional stability if desired.

A cannula-like extension was secured to the bottom of the adapter to provide additional support along the instrument shaft, reducing lateral movement. The extension was designed to match the effective length of da Vinci Si instruments by reducing the unsupported length of the X/Xi tool shaft. Interchangeable extensions accommodate both the 8 mm shafts of standard X/Xi instruments and the 12 mm shaft of the SureForm 45 stapler. Fig.~\ref{fig:instruments} shows the forceps, cautery hook, and stapler mounted on the same driver.
\subsection{Instrument Kinematics and Zeroing}

The da Vinci X/Xi instruments require coupling matrices to convert motor rotations into desired instrument joint movements. The fenestrated bipolar forceps, monopolar curved scissors, and large needle driver share a common coupling matrix shown below that map the shaft roll, wrist pitch, and grasper yaw and opening angle of the instrument to the rotation of the actuator disc shown in Figure \ref{fig:inners}.

\begin{equation}
\begin{bmatrix}
Disk\,1 \\ Disk\,2 \\ Disk\,4 \\Disk \,5
\end{bmatrix}
=
\begin{bmatrix}
t_r & 0 & 0 & 0 \\
0 & \lambda\,t_p & t_y & t_y/2 \\
0 & t_p & 0 & 0 \\
0 & -\lambda\,t_p & -t_y & t_y/2
\end{bmatrix}
\begin{bmatrix}
roll \\
pitch \\
yaw \\
grasp
\end{bmatrix}
\end{equation}
The coefficients $t_r,t_p,t_y$ are the transmission ratios for roll, pitch, and yaw, respectively. The $\lambda$ represents the amount of coupling between the disks generating pitch and yaw motions. The SureForm 45 stapler requires a separate coupling matrix to account for its different actuation mechanism, where there is shaft roll, wrist yaw, wrist pitch, and jaw closure.

These coefficients of the coupling matrices were calibrated experimentally by manually articulating the instrument end effector and measuring the joint angle and corresponding angular rotation of the instrument drive discs. The appropriate coupling matrix is selected based on the installed instrument, allowing this module to control different surgical tools.

Following instrument insertion, the spring-loaded drive discs must engage with the instrument discs. This can be accomplished manually with a joystick controller or by the external manual adjustment knobs. Alternatively, a homing script can be used to bring the instrument tip to a home configuration to establish initial joint positions and mechanical joint limits. For joints with detectable mechanical limits, a current-based homing routine rotates the motors in both directions and identifies the limits through increases in motor current. The midpoint between detected limits is used as the zero position. This was inspired by the da Vinci Xi auto-homing procedure. For joints without clear mechanical limits, the reference position is established through visual alignment, and can be refined manually.

\section{ENERGY INTERFACE}
\label{sec:energy}

\subsection{Design Goals}
\begin{figure}[t]
    \centering
    \begin{tikzpicture}[font=\sffamily\scriptsize, >={Stealth[length=1.8mm,width=1.4mm]},
        line cap=round, line join=round,
        esubox/.style={draw=black!80, line width=0.6pt, rounded corners=2pt, align=center,
            minimum width=13mm, minimum height=8mm, inner sep=1.5pt, fill=white},
        esupart/.style={esubox, minimum width=15mm, fill=black!7},
        esuoff/.style={draw=black!25, text=black!35, fill=white},
        esulab/.style={font=\sffamily\tiny, inner sep=1.5pt, align=center},
        esupath/.style={line width=1.1pt},
        esuidle/.style={line width=0.6pt, black!25}]
        \begin{scope}[yshift=3.35cm]
            \node[anchor=west, font=\sffamily\scriptsize\bfseries] at (-0.65,2.83) {(a) Board Unpowered};
            \node[anchor=east, font=\sffamily\tiny, text=black!60] at (6.05,2.8) {interposer board};
            \draw[black!55, line width=0.6pt, dash pattern=on 3pt off 2pt, rounded corners=3pt] (1.45,-0.05) rectangle (6.05,2.6);
            \node[esubox] (ped) at (0.03,1.95) {Foot\\Pedals};
            \node[esubox, esuoff] (pc) at (0.03,0.5) {Host PC\\(ROS~2)};
            \node[esupart] (byp) at (2.6,1.95) {Bypass\\Relays};
            \node[esupart, esuoff] (mcu) at (2.6,0.5) {ESP32};
            \node[esupart, esuoff] (act) at (4.95,0.5) {Activation\\Relays};
            \node[esubox] (esu) at (7.33,1.95) {ESU};
            \coordinate (j) at (4.95,1.95);
            \draw[esuidle] (pc) -- (mcu) (byp) -- (mcu) (mcu) -- (act) (act) -- (j);
            \draw[->, esupath, OIorange] (ped) -- (byp);
            \draw[->, esupath, OIorange] (byp) -- node[esulab, above, pos=0.4, text=OIorange!65!black]{direct path} (esu);
        \end{scope}
        \begin{scope}
            \node[anchor=west, font=\sffamily\scriptsize\bfseries] at (-0.65,2.83) {(b) Board Powered};
            \node[anchor=east, font=\sffamily\tiny, text=black!60] at (6.05,2.8) {interposer board};
            \draw[black!55, line width=0.6pt, dash pattern=on 3pt off 2pt, rounded corners=3pt] (1.45,-0.05) rectangle (6.05,2.6);
            \node[esubox] (ped) at (0.03,1.95) {Foot\\Pedals};
            \node[esubox] (pc) at (0.03,0.5) {Host PC\\(ROS~2)};
            \node[esupart] (byp) at (2.6,1.95) {Bypass\\Relays};
            \node[esupart] (mcu) at (2.6,0.5) {ESP32};
            \node[esupart] (act) at (4.95,0.5) {Activation\\Relays};
            \node[esubox] (esu) at (7.33,1.95) {ESU};
            \coordinate (j) at (4.95,1.95);
            \draw[esuidle, dash pattern=on 2pt off 2pt] (byp) -- node[esulab, above, text=black!45]{open} (j);
            \draw[->, esupath, OIblue] (ped) -- (byp);
            \draw[->, esupath, OIblue] (byp) -- node[esulab, left, pos=0.42, text=OIblue]{pedal\\sense} (mcu);
            \draw[->, esupath, OIblue] (mcu) -- node[esulab, above, text=OIblue]{drive} (act);
            \draw[esupath, OIblue] (act) -- node[esulab, left, text=OIblue]{contact\\closure} (j);
            \draw[->, esupath, OIblue] (j) -- (esu);
            \draw[<->, esupath, OIgreen] (pc) -- node[esulab, above, pos=0.42, text=OIgreen!65!black]{USB} (mcu);
        \end{scope}
    \end{tikzpicture}
    \caption{Signal paths through the ESU-TAP v2 interposer. (a) With the board unpowered, the bypass relays connect the foot pedals directly to the electrosurgical unit. (b) With the board powered, the controller processes pedal inputs and software commands and drives the activation relays.}
    \label{fig:esutap}
\end{figure}
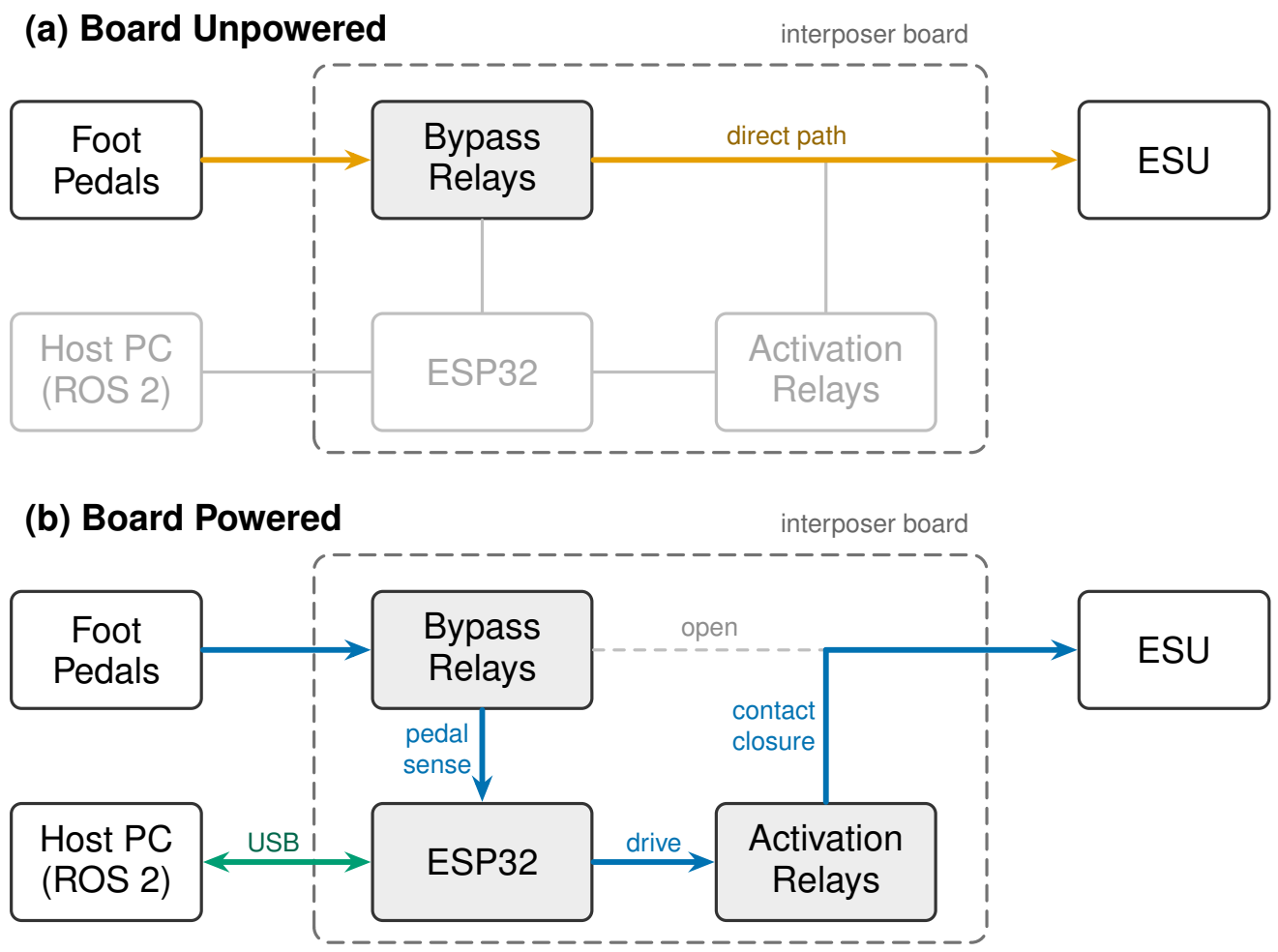

The energy interface provides software control and activation recording while retaining manual foot pedal operation. The ESU-TAP v2 interposer is a custom printed circuit board installed between the foot pedals and the electrosurgical unit (ESU), as shown in \Cref{fig:esutap}. The system uses Bovie BV-1253B monopolar and BV-1254B bipolar pedals with a DRE ASG-300 generator. The design focused on three objectives: (1) software activation of monopolar CUT, monopolar COAG, and bipolar modes, (2) a direct pedal path when unpowered, and (3) recording activation state and source. The interposer switches low-voltage footswitch contacts without modifying the generator or carrying its radiofrequency (RF) output. Other generators require a compatible footswitch interface and verification of wiring and signal levels.

\subsection{Interposer Circuit}
The interposer uses six Omron G6K-2F-Y DPDT relays: three activation relays, one per energy mode, and three bypass relays. One bypass relay switches the monopolar CUT and COAG lines, one switches their shared common, and one switches the bipolar line and its common. De-energized bypass relays connect the pedal lines and commons directly to the generator. When energized, they disconnect the pedal signal lines from the generator and connect the pedal commons to board ground for sensing. Activation relays then close the generator-side contacts for pedal or software requests. The pedal signal lines remain connected to sensing circuitry in both states. With the bypass relays energized, the generator-side contacts are separated from board ground; monopolar and bipolar generator commons remain on separate nets.

An ESP32-DevKitC-32E reads the pedals, drives activation relays, and communicates over USB serial using micro-ROS. Relays use a separate 5 V supply. The bypass coils share a return path switched by an AO3400A MOSFET under a TI TPS3823A watchdog supervisor. Firmware toggles the watchdog input approximately every 250 ms after a completed control cycle. A timeout briefly releases the bypass relays, without resetting the ESP32 or independently removing activation drive. The bypass provides manual fallback on loss of relay power, but does not guarantee uninterrupted activation during power restoration.

\subsection{Control and Pedal Priority}
A 10 ms control task samples pedals, applies commands, and writes relay outputs without ROS calls, allowing pedal forwarding without the host connection. A 30 ms debounce interval rejects brief input transitions, giving a nominal response delay of 30--40 ms before relay actuation. When an output opens, pending closures wait until at least the next control cycle. This break-before-make ordering is intended to prevent contact overlap during transitions. These timings are design values, not measured response guarantees.

Software control requires explicit enabling (arming) with the host connected and all pedals released. ON commands are renewed every 100 ms and expire 500 ms after the last accepted renewal. Expiry is checked by the control task and when processing commands, so a late renewal cannot revive an expired activation. Expiry or detected host disconnection clears software commands and disables software control. Any debounced pedal press while software control is enabled does the same in that control cycle; the pedal then controls the output subject to the interlock. Resuming software control requires deliberate re-enabling with all pedals released. These protections depend on the firmware remaining operational.

The board's ROS~2 interface is listed in \Cref{tab:esu-interface}. A robot-side host node exposes \texttt{/esu/enable} and accepts \texttt{/esu/target} requests at 10 Hz or faster for one software output at a time, renewing the board command every 100 ms. During teleoperation, software energy remains disabled and activation requests come from physical pedals.

\begin{table}[t]
\caption{ROS~2 interface of the ESU-TAP v2 interposer.}
\label{tab:esu-interface}
\centering
\footnotesize
\setlength{\tabcolsep}{3pt}
\begin{tabular}{@{}lll@{}}
\hline
Name & Type & Role \\
\hline
\texttt{/esu/arm} & \texttt{SetBool} & Enable/disable \\
\texttt{/esu/}\{\texttt{cut,coag,bipolar}\} & \texttt{SetBool} & Renew ON / set OFF \\
\texttt{/esu/activation\_state} & \texttt{Int32} & State and source \\
\texttt{/esu/capabilities} & \texttt{Trigger} & Board diagnostics \\
\hline
\end{tabular}
\end{table}

\subsection{Activation Recording}
Recording activation requests alongside robot motion provides action labels for learning from demonstration.
The interposer generates output-change events labelled by cause: pedal, software, or takeover. A takeover event is also emitted when control passes from software to a pedal without changing the output state. Relay commands and telemetry derive from the same controller state. Baseline events report current state on connection or resynchronization.

The host node republishes telemetry as \texttt{/esu/state} on changes and as 1 Hz snapshots. Teleoperation recordings use this topic's header timestamp for energy action labels. The board reports its firmware version, boot ID, and dropped-event count. Firmware-detected queue or publication failures increment this counter and trigger a baseline. Missed edges are not replayed, and these diagnostics cannot detect every loss in the recording pipeline.

The reported state represents commanded relay outputs, without contact or RF feedback. In the tested ASG-300, adding a second held pedal leaves only the first mode active; releasing the first while the second remains held produces fault F6 and inhibits output until both are released. Activation requests therefore do not establish delivered energy.

\section{EXPERIMENTS}
\label{sec:experiments}

\subsection{Experimental Design}

The electrosurgical interposer was evaluated using a DRE ASG-300 electrosurgical unit (ESU), configured to 10 W for monopolar CUT, monopolar COAG, and bipolar operation. All tests were performed without surgical instruments connected to the generator outputs. Each activation mode was tested using both software commands and physical foot pedals, with generator activation verified through its audible tones and visual indicators. Pedal activation was also tested with the interposer unpowered to verify the hardware bypass mechanism.

The software activation timeout was evaluated over eight trials by issuing activation commands renewed every 100 ms, then stopping the renewals without sending a deactivation command. Pedal takeover was tested by pressing the bipolar foot pedal during software-controlled CUT activation. Communication loss was simulated by terminating the micro-ROS agent during an active software command. Finally, activation-state recording was evaluated through six foot pedal presses across the three energy modes.

Three ex vivo tasks evaluated the instrument driver and energy interface together on ARTS, an open-source three-arm research platform~\cite{ge2026arts}. In the first task, a suction tool developed at Johns Hopkins University and described by Fan et al.~\cite{fan2026supervisedautonomousbimanualsuction} was used to clear simulated blood from a splenic wound before monopolar coagulation was applied at the wound site. The second task used bipolar forceps to approximate and coagulate splenic wound edges. The third used monopolar cutting during gallbladder dissection. Video, robot kinematics, and electrosurgical activation events were recorded during each task.

Instrument exchange time was measured over 20 trials using the protocol described for the DASH Xi adapter~\cite{wickenhiser2026dash}, alternating between the Fenestrated Bipolar Forceps and Monopolar Curved Scissors. The UR5’s built-in payload estimation function was used to estimate module mass alone and with each instrument listed in Table~\ref{tab:mass}. Mass was excluded from the direct comparison with DASH because the two evaluations used different instrument sets.

The kinematic accuracy of the instrument driver was evaluated using the Fenestrated Bipolar Forceps and its calibrated coupling matrix. An optical tracking marker was attached to the forceps, and the adapter was commanded to rotate the marker about the roll, pitch, and yaw axes in $5^\circ$ increments. The marker pose was recorded at each increment.

Homing repeatability was evaluated over ten trials with the Permanent Cautery Hook. Its roll (M1), pitch (M4), and yaw (M5) axes have detectable mechanical limits, allowing current-based homing without visual jaw alignment. The final encoder position for each axis and total homing time were recorded. The hook was removed and reinstalled between successive trials.

\subsection{Evaluation Metrics}

The timeout interval was measured from the final acknowledged software command to the board's reported deactivation. This measurement evaluates the ability of the interposer to automatically disable software-controlled activation when commands are no longer renewed.

Pedal takeover was evaluated by measuring the interval between deactivation of the software-controlled output and activation of the pedal-controlled output. Communication loss was evaluated by observing generator activation and the interposer's reported state following termination of the micro-ROS agent. These tests assess the ability to return control to the physical pedals and terminate software activation when communication is interrupted. The number and order of recorded activation and deactivation events were compared with the physical foot pedal inputs. This measurement evaluates the reliability of recording electrosurgical activation events for subsequent analysis of surgical procedures.

Video review assessed three observable outcomes: clearance of simulated blood and application of monopolar coagulation at the wound site, persistence of wound-edge approximation after release of the bipolar forceps, and tissue separation during monopolar dissection.

\begin{figure}[!t]
    \centering
    \includegraphics[width=0.8\columnwidth]{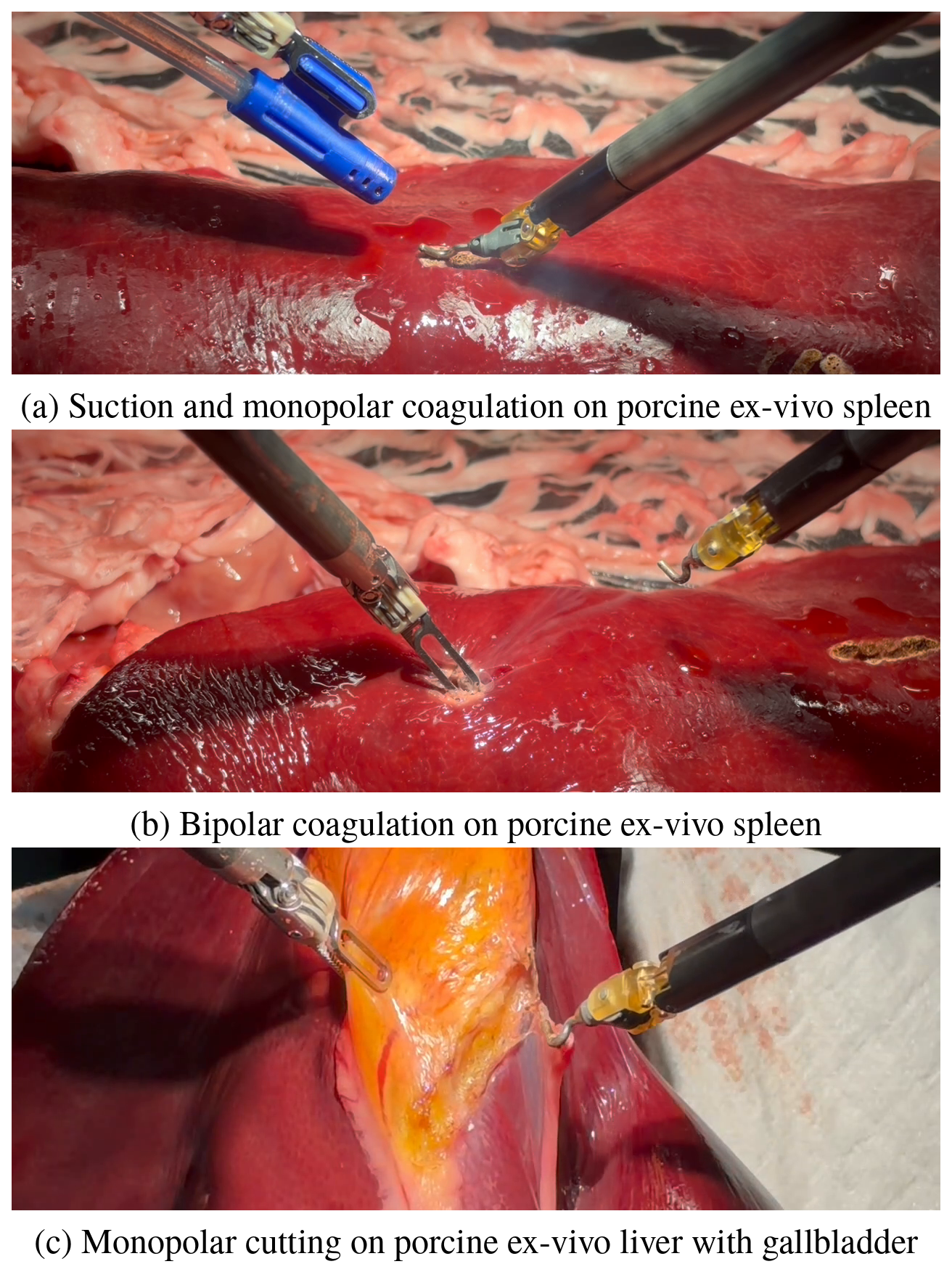}
    \caption{Ex vivo electrosurgical tasks performed with the instrument module. (a) Suction and monopolar coagulation at a splenic wound. (b) Bipolar coagulation of approximated splenic wound edges. (c) Monopolar cutting during gallbladder dissection.}
    \label{fig:exvivo}
\end{figure}

Following DASH~\cite{wickenhiser2026dash}, instrument exchange time was measured from first contact with the installed instrument to full seating of the replacement, excluding automatic homing. Module mass was reported with and without an instrument attached to characterize the payload on the robotic arm.

For each measured pose of the optical marker, the axis and angle of rotation were calculated relative to the initial pose at zero joint angles. Measured rotation angles were compared with commanded angles to evaluate the transmission ratios. The angles between the measured roll, pitch, and yaw axes were used to evaluate the coupling coefficients.

Homing repeatability was quantified for each axis using the standard deviation and range of the final encoder positions across trials. Both measures were converted to equivalent joint angles using the encoder resolution of 0.088° per tick and the axis-specific transmission ratio.

\section{RESULTS}
\label{sec:results}

The interposer successfully activated monopolar CUT, monopolar COAG, and bipolar modes through both software commands and physical foot pedals, as confirmed by the ESU's audible tones and visual indicators. All three modes remained operable through the physical pedals with the interposer unpowered, confirming the functionality of the hardware bypass.

The board-reported software activation timeout was $503.1 \pm 3.8$ ms across eight trials, consistent with the programmed 500 ms timeout. During pedal takeover, the board reported software-controlled CUT released and bipolar closed 11 ms later. Subsequent software commands were rejected until software control was explicitly re-enabled. When communication was interrupted by terminating the micro-ROS agent, the ESU ceased activation without additional commands. Upon reconnection, the interposer reported software control disabled and all outputs deactivated.

The interposer also successfully recorded all 12 activation and deactivation events across six foot pedal presses, preserving the correct sequence and identifying the physical pedals as the activation source. The timeout and takeover times are board-reported; contact opening and RF output were not timed independently, and no instrument was connected, so these tests establish interface behavior rather than tissue performance.

The accompanying videos show suction clearing simulated blood from a splenic wound and monopolar coagulation at the wound site. Fig.~\ref{fig:exvivo} shows representative images from the three ex vivo electrosurgical tasks. Following bipolar coagulation, the splenic wound edges remained together after the forceps jaws opened. The cholecystectomy video shows tissue separation during monopolar gallbladder dissection.

Across 20 trials, instrument exchange time was $2.41 \pm 0.34$~s (mean $\pm$ SD). DASH Xi reported an exchange time of $3.39 \pm 0.51$~s~\cite{wickenhiser2026dash}. The estimated module mass ranged from 0.740~kg without an instrument to 1.270~kg with the SureForm 45 Stapler (Table~\ref{tab:mass}).

\begin{table}[h]
\centering
\caption{Mass of the instrument module with different da Vinci instruments.}
\label{tab:mass}
\begin{tabular}{@{}lc@{}}
\toprule
Configuration & Mass (kg) \\
\midrule
Module only                       & 0.740 \\
Large needle driver               & 0.850 \\
Fenestrated Bipolar Forceps               & 0.830 \\
Monopolar Curved Scissors               & 1.030 \\
Permanent Cautery Hook               & 0.960 \\
SureForm 45 Stapler               & 1.270 \\
\bottomrule
\end{tabular}
\end{table}

\begin{figure}[!t]
    \centering
    \includegraphics[width=\columnwidth]{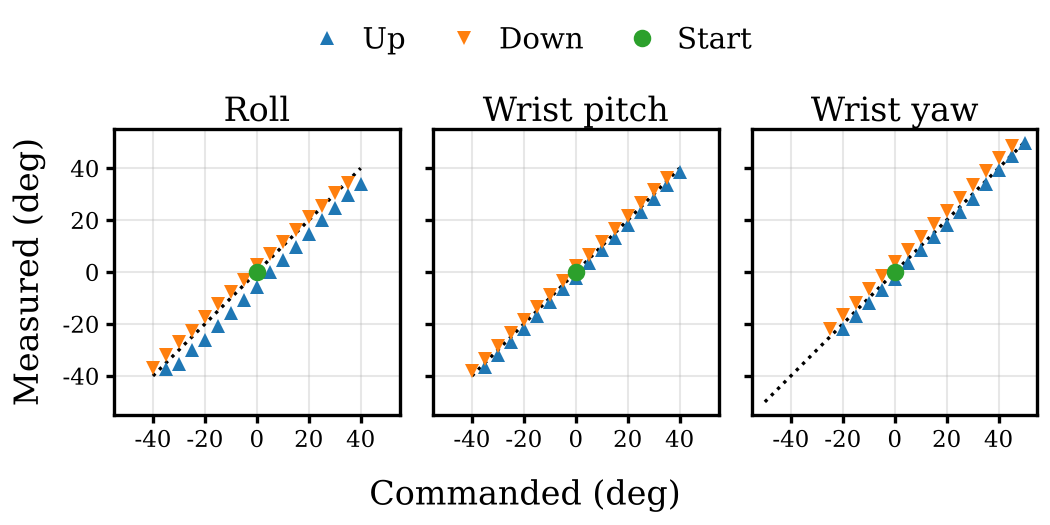}
    \caption{Commanded and optically measured rotation angles for roll, pitch, and yaw of the Fenestrated Bipolar Forceps after refinement of the coupling matrix.}
    \label{fig:coupling}
\end{figure}

Initial coefficients of the coupling matrix of the bipolar grasper obtained from the manual calibration were $[t_r, t_p, t_y, \lambda] = [1.0, 0.95, 1.28, 0.63]$. The marker was rotated from $-40^\circ$ to $40^\circ$ in roll and pitch axis, $-25^\circ$ to $40^\circ$ in yaw axis. The ratio of measured to commanded rotation angle, which will be close to 1 if the transmission ratios were accurate, was $[0.993, 0.904, 1.14]$ in roll, pitch and yaw rotation. The angle between roll-yaw and pitch-yaw axes, which indicates coupling between the rotations were $80.1^\circ$ and $87.9^\circ$, respectively. Using the measurement data from the initial experiments, coefficients were fine tuned and the experiment was repeated using the fine tuned coupling matrix. The fine tuned coefficients were
$[t_r, t_p, t_y, \lambda] = [1.0, 1.050, 1.12, 0.57]$.After calibration, the measured-to-commanded rotation ratios were 0.99, 1.00, and 1.02 for roll, pitch, and yaw, respectively, showing improvement in pitch and yaw direction.
The angle between roll-yaw and pitch-yaw axes improved to  $89.85^\circ$ and $89.3^\circ$, respectively. Figure~\ref{fig:coupling} shows the plot between the commanded and actual rotation angle in roll, pitch and yaw axes using the fine tuned coupling matrix. While the slope of the plot is close to 1, it clearly shows the hysteresis, which are inherent in the cable driven mechanisms.

Mean homing time was 34.9 ± 0.2 s across ten trials. Pitch had the smallest spread in homed position, with a range of 9 ticks (0.84° equivalent joint angle). Yaw had the largest spread, with a range of 14 ticks (1.12$^\circ$ equivalent joint angle; Table~\ref{tab:hook_homing_repeatability}).
\begin{table}[!t]
    \centering
    \caption{Cautery hook homing repeatability over 10 trials. SD denotes standard deviation, and range is the difference between the maximum and minimum homed positions. Equivalent joint angles were calculated from motor encoder readings.}
    \label{tab:hook_homing_repeatability}
    \footnotesize
    \begin{tabular}{@{}lrrrr@{}}
        \hline
        & \multicolumn{2}{c}{Encoder (ticks)}
        & \multicolumn{2}{c}{Equivalent joint angle ($^\circ$)} \\
        \cline{2-3} \cline{4-5}
        Axis & SD & Range & SD & Range \\
        \hline
        Roll (M1)  & 3.1 & 12 & 0.25 & 0.99 \\
        Pitch (M4) & 2.6 &  9 & 0.24 & 0.84 \\
        Yaw (M5)   & 4.4 & 14 & 0.36 & 1.12 \\
        \hline
    \end{tabular}
\end{table}

\section{DISCUSSION}
\label{sec:discussion}

The instrument module allows researchers to exchange da Vinci X/Xi instruments on a UR5/UR5e robotic arm and record electrosurgical activation. The ex vivo videos showed tissue separation during monopolar cutting. After bipolar coagulation, the splenic wound edges remained together when the forceps opened. The simulated bleeding task demonstrated suction and monopolar coagulation at the wound site.

The mechanical evaluation showed the importance of calibrating the driver for each instrument. For the Fenestrated Bipolar Forceps, calibration brought the measured pitch and yaw rotations closer to their commanded values in both directions. Across 20 tool exchanges, the mean time to replace an instrument was 2.41 seconds. The time required to resume instrument control also includes homing.

For the SureForm 45 Stapler, we determined the coupling matrix for shaft roll, wrist articulation, and jaw motion. Staple firing remains a separate development step, and future work will evaluate deployment using the instrument module. Unlike DASH~\cite{wickenhiser2026dash}, which uses RFID identification to load instrument-specific kinematics and perform zeroing, this module requires manual selection of the instrument configuration. For the energy interface, the interposer's firmware lease and pedal takeover were verified. A hung controller or a welded relay is not covered by the board and is handled by the generator's power switch.

\section{CONCLUSION}
\label{sec:conclusion}
The open-source ARTS instrument module combines a five-motor driver for da Vinci X/Xi instruments with an electrosurgical interface that preserves pedal priority. Across ten cautery-hook trials, current-based homing established motor references with encoder-derived standard deviations equivalent to 0.24°–0.36° at the joints. The interface recorded all 12 pedal transitions in bench testing, and ex vivo demonstrations paired activation records with robot motion and video during cutting and coagulation. The driver also accommodates the SureForm 45 Stapler, with a coupling matrix for shaft roll, wrist articulation, and jaw motion. Future work will evaluate staple firing. CAD files, firmware, and driver software are available through the ARTS repository.

\bibliography{bibl}
\bibliographystyle{IEEEtran}

\end{document}